\documentclass[11pt]{article} % For LaTeX2e
\usepackage{rldmsubmit,palatino}
\usepackage{graphicx}

\usepackage{natbib}
\usepackage{enumitem}
\usepackage{wrapfig}
\usepackage{algorithm}
\usepackage{algpseudocode}
\usepackage{hyperref}
\usepackage{amssymb}
\usepackage{xparse}

\let\emptyset\varnothing

\title{Better Decisions through the Right Causal World Model}

\author{
Elisabeth Dillies\thanks{Equal contribution} \\
Department of Cognitive Science\\
Sorbonne University\\
\texttt{elisabeth.dillies@gmail.com} 
\And
Quentin Delfosse$^{*}$ \\
Department of Computer Science\\
Technical University Darmstadt \\
\texttt{quentin.delfosse@tu-darmstadt.de} \\
\AND
Jannis Blüml \\
Department of Computer Science\\
Technical University Darmstadt \\
\And
Raban Emunds \\
Department of Computer Science\\
Technical University Darmstadt \\
\AND
Florian Peter Busch \\
Department of Computer Science\\
Technical University Darmstadt \\
\And
Kristian Kersting \\
Department of Computer Science\\
Technical University Darmstadt \\
}

\newcommand{\eg}{\emph{e.g.}~} 

\newcommand{\ie}{\emph{i.e.}~}

\NewDocumentCommand{\ram}{o}{\texttt{ram[#1]}}

\begin{document}

\maketitle

\begin{abstract}
Reinforcement learning (RL) agents have shown remarkable performances in various environments, where they can discover effective policies directly from sensory inputs. However, these agents often exploit spurious correlations in the training data, resulting in brittle behaviours that fail to generalize to new or slightly modified environments. 
To address this, we introduce the Causal Object-centric Model Extraction Tool (COMET), a novel algorithm designed to learn the exact interpretable causal world models (CWMs). COMET first extracts object-centric state descriptions from observations and identifies the environment’s internal states related to the depicted objects' properties. Using symbolic regression, it models object-centric transitions and derives causal relationships governing object dynamics. COMET further incorporates large language models (LLMs) for semantic inference, annotating causal variables to enhance interpretability.

By leveraging these capabilities, COMET constructs CWMs that align with the true causal structure of the environment, enabling agents to focus on task-relevant features. The extracted CWMs mitigate the danger of shortcuts, permitting the development of RL systems capable of better planning and decision-making across dynamic scenarios. 
Our results, validated in Atari environments such as Pong and Freeway, demonstrate the accuracy and robustness of COMET, highlighting its potential to bridge the gap between object-centric reasoning and causal inference in reinforcement learning.
\end{abstract}

\keywords{
reinforcement learning, world model, object-centric, causality
}

\acknowledgements{This research work has been funded by the German Federal Ministry of Education and Research, the Hessian Ministry of Higher Education, Research, Science and the Arts (HMWK) within their joint support of the National Research Center for Applied Cybersecurity ATHENE, via the ``SenPai: XReLeaS'' project as well as their cluster project within the Hessian Center for AI (hessian.AI) ``The Third Wave of Artificial Intelligence - 3AI''.}

\startmain % to start the main 1-4 pages of the submission.

\section{Introduction}

% \textbf{Deep RL agents can solve a wide variety of tasks} ATARI. 

% \begin{wrapfigure}[13]{r}{.34\textwidth}
%     \centering
%     \vspace{-4mm}
%     \includegraphics[width=1.\linewidth]{exampleframes.pdf} 
%     %\vspace{10mm}
%     \caption{Pong and Freeway states.}
%     \label{fig:da_not_robust}    
% \end{wrapfigure}

\begin{figure}[b]
    \includegraphics[width=1.\linewidth]{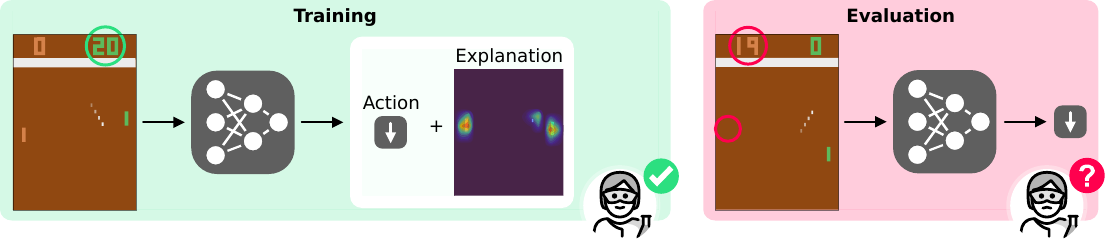} 
    \caption{\textbf{Deep RL agents learn undetectable shortcut.} A deep (PPO) agent reaches the maximum score and selects the correct action in the depicted state of the training environment. The explanation map further leads external reviewers to think that the agent ``understands'' that it should return the ball behind the enemy (Left). In the test environment, where the enemy is hidden, the agent decides to go down, preventing it from catching the ball (Right). This illustrates that the agents learned to rely on the enemy's vertical position as an estimation of the ball's position. %Borrowed from~\citep{delfosse2024interpretable}.
    }
    \label{fig:motivation}    
\end{figure}

Machine learning systems often tend to exploit spurious correlations in the training data~\citep{SchramowskiSTBH20,rightconcepts}. This is no different in RL settings, where agents learn misaligned policies that fail to learn the originally intended tasks by relying on shortcuts~\citep{Langosco2022goal, SuauSO24}.
Such shortcut learning even occurs in the most simple Pong Atari game, for which~\citet{delfosse2024interpretable} showed that both deep and symbolic RL agents are subject to misalignment. 
In Pong, the enemy is programmed to vertically align itself with the ball, leading to a quasi perfect correlation between the two objects vertical positions. 
RL agents thus learn to rely on the vertical position of the enemy to situate the ball position and effectively return it, as depicted in Figure~\ref{fig:motivation}.
These shortcuts prevent the agents from generalizing, even to simplified versions of the environments, for which humans have no problem adapting. 
For example, if the enemy is hidden, or stops moving when the ball is going towards the player, the RL agents' performances drop. These misgeneralization to simpler scenarios are not specific to Pong, but are recurrent throughout the arcade learning environments~\citep{delfosse2024hackatari}. 

% \textbf{Object centric agents allow for detection+correction.}
Many form of interpretable object-centric algorithms have recently been developed. They first extract object-centric (or symbolic) states, then rely on first order logic~\citep{Delfosse2023InterpretableAE}, polynomial approximations and LLM explanations~\citep{insight2024} or decision trees~\citep{kohler2024interpretable}. They have been shown to be competitive alternatives to opaque approaches, performing on par with deep agents. Importantly, they provide some form of interpretability, that allow experts to detect and correct their potential misalignment behaviors. \citet{yoon2023investigation} demonstrate the higher robustness of object-centric agents in improving generalization and reducing reliance on spurious correlations.

However, object-centric symbolic reasoning does not break up correlations like the one between the enemy's paddle and the ball in Pong.
If the enemy starts behaving differently, an agent relying on this spurious correlation and focusing on the enemy paddle will fail to return the ball, while an agent moving based on the ball position and velocity will not be affected by the correlation break.
The difference between the relevance of both observable features (\ie the enemy paddle and ball vertical positions) for our intended agent behavior lies in the true causal relations.
%Whether the enemy scores is directly determined by the position of the agent's paddle and ball but not by the enemy's paddle.
If the agent utilizes these causal relationships in a causal world model, the independence of such mechanisms~\citep{peters2017elements} ensures that the desired behaviour (\ie returning the ball) applies in modified environments (\eg with an enemy stopping after returning the ball).
By disentangling these relationships, the agent can focus on task-relevant features, fostering the development of policies that generalize across diverse scenarios, including those with novel dynamics or adversarial interventions.

Even if novel benchmarks for exposing such misgeneralizations have been developed to test RL agents' robustness, (\eg ~\citep{delfosse2024hackatari} in the Atari domain), methods to automatically detect and correct spurious correlations are still underexplored. 
Integrating causal reasoning into RL agents offers a pathway toward human-like adaptability~\cite{yang2024towards,lei2024spartan}. By abstracting relevant features and modeling the causal structures underlying the observed phenomena, RL agents could autonomously overcome the limitations of shortcut-driven strategies. 

In this paper, we introduce the \textbf{C}ausal \textbf{O}bject-centric \textbf{M}odel \textbf{E}xtraction \textbf{T}ool (COMET). 
Using their internal states, COMET aims to extract the interpretable causal world models (CWMs) of simulated environments. COMET detects the objects from the observation and then models causal relationships between the depicted objects' properties and the internal state (\eg the RAM) that the emulated environment relies on to produce the observations. It then learns to extract \textit{when} and \textit{how} these variables are updated by the environment. 
It retrieves the relevant internal variables and uses the common reasoning abilities of an LLM to annotate these relevant internal states (thus improving its interpretability).

\newpage

\section{Method}

The algorithm, detailed in Algorithm~\ref{alg:cap}, outlines the process for extracting a causal world model (CWM) using COMET. This method integrates object-centric reasoning with symbolic regression and semantic inference to map environmental dynamics comprehensively. Below, we describe each stage of the extraction process.

Overall, COMET extracts object-centric CWM by:
\begin{enumerate}%[itemsep=2pt,parsep=1pt,topsep=0pt,partopsep=2pt]
    \item Mapping the internal state values to the objects' properties,
    \item Modeling how this internal state evolves,
    \item Annotating the relevant variables with their meaning.
\end{enumerate}

COMET requires an executable environment, an executable policy (potentially random) and a large language model (LLM) for common sense reasoning.
It first generates rollouts from the policy, retrieving the environment's internal states (EIS) and rendered RGB observations. 
It then uses an object discovery method (such as ~\citep{Delfosse2021MOC, 
zhao2023fast}) to extract the objects from each frame. 
Each object consists of a set of properties (such as the $\texttt{x}, \texttt{y}, \texttt{w}, \texttt{h}$ bounding box coordinates or the \texttt{value} of objects such as scores, lives counter, etc.).

COMET then performs symbolic regressions, using PySR~\citep{cranmer2023interpretable}, to match each property with their corresponding internal states. 
It thus extracts a subset, R-EIS, of relevant internal states and their mapping to the different objects' properties. For example, the internal state $\texttt{s}_i$ can encode the observed vertical position of the depicted ball object, with an offset of $14$, thus following: $\texttt{Ball.y} = \texttt{s}_i - 14$. 

To extract the underlying object-centric world model, \ie understand how the objects evolve, COMET then searches both \textit{if} and \textit{how} these relevant internal states are updated at each step. 
Symbolic regressions are performed to map each already collected relevant state to internal states and actions. 
For example, our symbolic regression model can extract the following mapping: $\texttt{s}_i = \texttt{s}_i + \texttt{s}_j$, with $s_j$ another internal state, that here corresponds to the vertical speed of the ball. 
If new relevant states are found within the equation (\eg $s_j$ here), they are added to R-EIS. 
This relevant state regression is repeated until COMET has obtained the update conditions and equations of all the relevant states (which eventually happens, as the set of the environment's internal states is finite).

We then aim to improve the interpretability of the extracted world model, by annotating the relevant internal states with their meaning. To identify the semantics behind $s_j$, we use the common sense reasoning abilities of an LLM. 
Specifically, we provide ChatGPT (model 4o) with the equations that lead back to the object properties and ask for an annotation of the internal states' semantics. In the situation described above, the LLM correctly identifies\footnote{\href{https://chatgpt.com/share/6786f2ef-3ab4-8006-b5e3-8a3b29e92b2e}{https://chatgpt.com/share/6786f2ef-3ab4-8006-b5e3-8a3b29e92b2e}, last accessed on 15.01.2024} that $s_j$ encodes the vertical velocity of the Ball. 
The provided semantics allow external reviewers to better identify each internal state purpose. 
Further, it can allow the LLM to reduce the set of internal state variables on which the regression is done. 
For example, the Ball's horizontal velocity in Pong is flipped when the ball bounces on one of the paddle. 
An LLM can detect that such event happens when the ball is colliding with another object, and reduce the input set to the other objects positions only. 
Even if facing a novel situation, for which the LLM might not know the task, a priori, its common sense reasoning can still lead it to identify that a sudden change in the a traveling object's velocity is due to a collision, and thus direct the search towards collision detection.

%\begin{minipage}{.75\textwidth}
\begin{algorithm}%[H]
\caption{Causal Object-centric Model Extraction}
\label{alg:cap}
\begin{algorithmic}[1]
\Require env, agent, LLM
\State \textbf{init} worldmodel
\State rgbs, EIS, actions $\gets$ sample(env, agent, nb\_episodes)
\State objs $\gets$ detect(rgbs)
\State R-EIS $\gets$ find\_relevant\_EIS(objs.properties, EIS)
\While{R-EIS $\neq \emptyset$}
    \State \texttt{s} $\gets$ R-EIS.pop()
    \If {\texttt{s} $\notin$ worldmodel}
        \State update\_equation, update\_condition $\gets$ find\_hidden\_state(EIS, actions)
        \State worldmodel(\texttt{s}) $\gets$ update\_equation, update\_condition
        \State R-EIS $\gets$ R-EIS + update\_equation.variables
    \EndIf
\EndWhile
\State annotate\_variables(worldmodel, LLM)
\State \Return worldmodel
\end{algorithmic}
\end{algorithm}
%\end{minipage}
% \begin{minipage}{.2\textwidth}
% text
% \end{minipage}

 \newpage

\begin{figure}[t]
    \centering
    \includegraphics[width=0.999\linewidth]{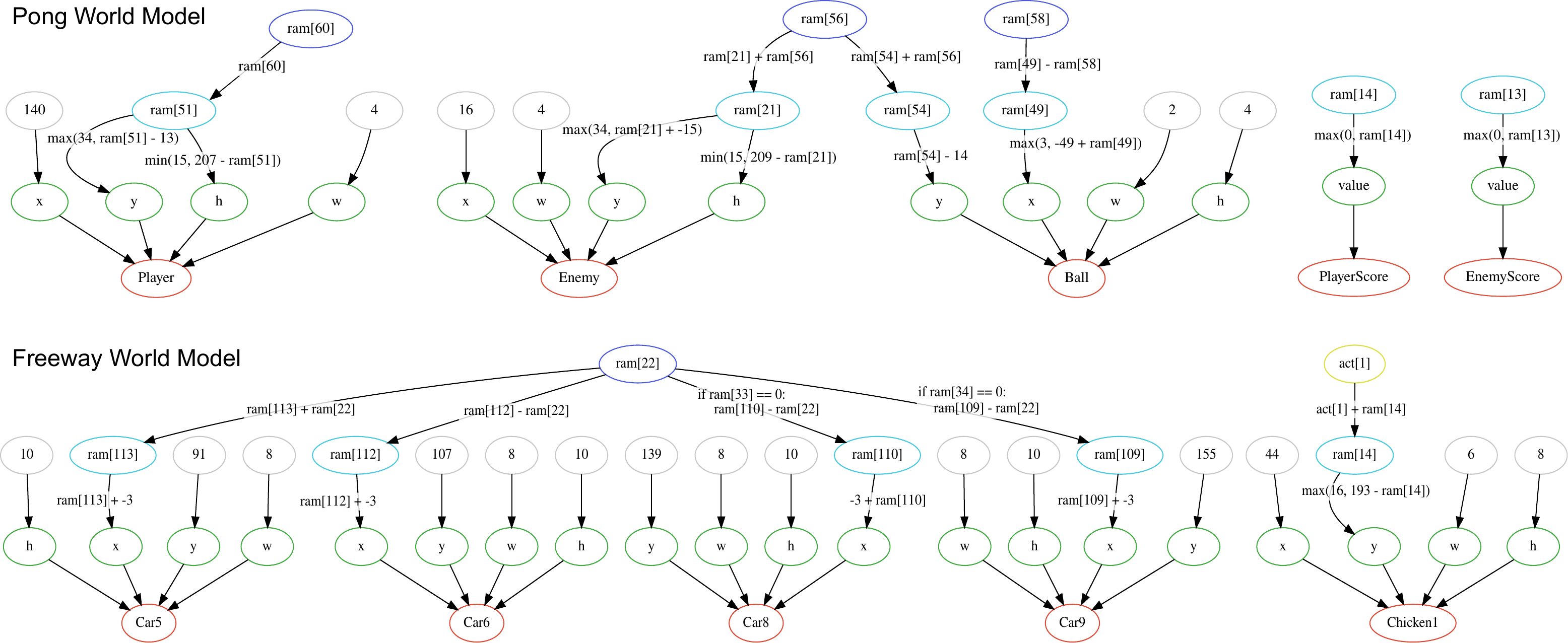}
    \caption{\textbf{Example of world models extracted by COMET.} Top: The extracted world model for the Pong environment, that consist of the player, the enemy, their scores, and the Ball. Bottom: Part of the Freeway world model ($4$ depicted cars out of $10$). 
    All the internal RAM variables extracted directly linked to the objects' properties (light blue) are on par with the annotated RAM provided by \citet{anand2019unsupervised}. The variables used to update these properties (dark blue) are valid. For example, \ram[58] corresponds to the speed of the Ball on the x-axis.}
    \label{fig:wms}
\end{figure}

\section{COMETs extracted world models}

This section showcases two extracted world models obtained from the Pong and Freeway environments. We use the emulator random access memory (RAM) as internal states of the environments.
For these, we use the object extractor incorporated in OCAtari~\citep{Delfosse2024OCAtari} and sample transitions by letting a human agent play for $1$ game episode (approximately $8000$ steps). To identify if COMET finds the correct RAM values, we first refer to the Atari RAM annotations provided by \citet{anand2019unsupervised}\footnote{\href{https://github.com/mila-iqia/atari-representation-learning/blob/master/atariari/benchmark/ram_annotations.py}{https://github.com/mila-iqia/atari-representation-learning/blob/master/atariari/benchmark/ram\_annotations.py}}. 
To assert if the other variables (from the update conditions) are correct, we directly alter them and check if they lead to the expected depicted modification. For example, we can set the horizontal speed of the ball to $0$ and observe it moving vertically.

As depicted in Figure~\ref{fig:wms}, COMET has correctly identified the properties' states for both games. 
For Pong, COMET correctly extracts the horizontal and vertical speed of the ball (\ie \ram[58] and \ram[56] respectively). 
The state \texttt{\ram[51]} is indeed updated to become \ram[60] (updated based on the agent's selected action). However, \ram[56] has also been retrieved to encode the player's speed. This is because the enemy is programmed to follow the ball (as detailed in the introduction). The enemy's speed, thus, indeed matches the ball's in the vast majority of the game transition.
However, the enemy is programmed to catch the ball and thus follows it. If the ball is below the enemy, the enemy will decide to go down. However, this rule is more complicated than the one matching the enemy's speed. Thus, PySR favours the simpler rule. Two solutions can be applied here: allow the model to perform intervention, \ie modify the speed of the ball to check if that leads to a direct modification of the enemy's y position update rule, or use the common sense reasoning of an LLM to select the most appropriate update rule. 

In Freeway, the agent controls \texttt{Chicken1}, aiming to cross the road without getting hit by cars that move horizontally. The chicken's vertical position is updated based on the agent's selected action. It is incremented based on the direction of the joystick (\ie act[1]). The different cars' positions are incremented for cars $1$ to $5$ and decremented for cars $6$ to $10$.
However, to simulate the different speed for each car, the car \texttt{x} positions are updated at different paces. While the \texttt{x} positions of car5 and car6 get updated at each step, other cars like car8 and car9 use counters to be updated every $3$ and $4$ frames, respectively. 
The update conditions of car8 and car9 horizontal positions are thus correctly identified by COMET.
Finally, in the regressed set \ram[22] is constantly set to $-1$ (only set to $0$ when the game is over, a state for which the cars' positions are indeed not updated). Altering \ram[22] does not affect the speed of the car. 
While the regression leads to a correct result, an intervention altering the internal state value would allow COMET to identify and correct its mistake. 
The LLM's reasoning could also here detect that \ram[22] represents the game being over (or not) from the regressions' inputs. 
A simpler alternative is to punish the symbolic regressor for using variable instead of constant.

\section{Discussion and future work}
\label{sec:discussion}

In this paper, we introduced COMET and an algorithm to extract an interpretable object-centric causal world model by performing symbolic regression on the observable objects and providing semantics to the causal variables by leveraging the common sense reasoning capabilities of LLMs.
One of the major discussion points of COMET is the fact that it accesses the hidden state of the environment. 
Most methods that extract CWM for RL agents are not given access to the internal states~\citep{yang2024towards, lei2024spartan}, which constitute more realistic settings. 
However, most RL agents are (at least) pretrained learning within virtual environments, even in industrial settings. 
COMET aims to extract the exact CWM from the environment, which can serve as the target CWM. 
Accessing the true CWM allows us to reimplement these exact environments using JAX.\footnote{\href{https://github.com/k4ntz/JAXAtari}{https://github.com/k4ntz/JAXAtari}}. Initial benchmark of our JAX version of \eg Pong, using GPU-based parallelization (on an RTX2070) lead to speedups between $30$ to $100$, compared to the CPU execution of the original Atari gym version. %Specifically, it emulates around $1.2$ million steps per second (using $1000$ environment workers) compared to only $67000$ steps per second for the Atari baseline achieves on $16$ CPU cores (9-9900KF CPU @ 3.60GHz).

Our most important next step is incorporating the LLM's common reasoning abilities within COMET. We will use the LLM to generate symbolic functions (in julia and sympy), that can be used by PySR at regression time.
We have made interventions on the internal relevant variables extracted by COMET to test the accuracy of the extracted model. We plan to integrate the interventions within COMET, to allow for correcting the CWM.
Finally, we, of course, plan to extend our evaluations to more environments, notably from the ALE suite.

\bibliographystyle{plainnat}
\bibliography{bibliography}

% \section{\textit{Some relevant Causality papers}}
% \begin{itemize}
%     % \item \cite{sauter2024core}: Agent picks intervention to help with causal discovery
    
%     % \item \cite{zhucausal}:  ``Classical'' causal discovery setting where RL is used as a method (not as a setting)
%     % \item \cite{zhang2019near}: Something on treatment estimation using RL (need to take a closer look if we are interested in details)
%     % \item \cite{ruan2023causal}: Imitation learning for causal relations and with the possibility of confounding
%     % \item \cite{pearl2009causality}: Nothing RL, but the default causality / SCM reference
%     % \item \cite{zeng2024survey}: Haven't read it, but it is a very recent Causal RL survey, might include that
%     % \item \cite{madumal2020explainable}: Learn SCMs in RL setting, mostly to generate (counterfactual) explanations
%     % \item \cite{seitzer2021causal}: Looking for (``situation-dependant``) causal relations in RL to improve RL (exploiting actions with high causal influence, changing rewards based on causality, ...)
%     % \item \cite{yu2024adam}: focus on lifelong-learning, agent makes interventions to find causal graphs; pretty elaborate experimental setup with Minecraft

%     \item TO BE INCLUDED: 
%     \item \cite{lei2024spartan}: Causal graphs with attention (i.e., functions between variables may or may not be present)
%     \item \cite{yang2024towards}: Causality + RL, they also consider changes in the environment; similar kinds of experiments as we plan to have
% \end{itemize}

\end{document}